\documentclass[sigconf, authorversion]{acmart}

\pdfoutput=1
\usepackage{booktabs} % For formal tables
\usepackage[utf8]{inputenc}
\usepackage[english]{babel}
\usepackage[LAE,T1]{fontenc}
\usepackage{booktabs}
\usepackage{multirow}
\newcommand{\arabictext}[1]{\bgroup\beginR\fontencoding{LAE}\selectfont #1 \endR\egroup}

\setcopyright{acmlicensed}
\acmDOI{10.1145/3605098.3636065}

\acmISBN{979-8-4007-0243-3/24/04}

\acmConference[SAC'24]{ACM SAC Conference}{April 8 –April 12, 2024}{Avila, Spain}
\acmYear{2024}
\copyrightyear{2024}

\acmArticle{4}
\acmPrice{15.00}

\begin{document}
\title{Ar-Spider: Text-to-SQL in Arabic}
%\titlenote{Produces the permission block, and
%  copyright information}
%\subtitle{Extended Abstract}
%\subtitlenote{The full version of the author's guide is available as
%  \texttt{acmart.pdf} document}
  
\renewcommand{\shorttitle}{Ar-Spider, Arabic text-to-SQL dataset}

\author{Saleh Almohaimeed}
\orcid{1234-5678-9012}
\affiliation{%
  \institution{University of Central Florida}
  %\streetaddress{P.O. Box 1212}
  \city{Orlando} 
  \state{Florida} 
  \country{USA}
}
\email{sa247216@ucf.edu}

\author{Saad Almohaimeed}
\affiliation{%
  \institution{University of Central Florida}
  %\streetaddress{P.O. Box 1212}
  \city{Orlando} 
  \state{Florida} 
  \country{USA}
}
\email{sa583575@ucf.edu}

\author{Mansour Al Ghanim}
\affiliation{%
  \institution{University of Central Florida}
  %\streetaddress{P.O. Box 1212}
  \city{Orlando} 
  \state{Florida} 
  \country{USA}
}
\email{mansour.alghanim@ucf.edu}

\author{Liqiang Wang}
\affiliation{%
  \institution{University of Central Florida}
  %\streetaddress{P.O. Box 1212}
  \city{Orlando} 
  \state{Florida} 
  \country{USA}
}
\email{lwang@cs.ucf.edu}

\renewcommand{\shortauthors}{Sa Almohaimeed.}

\begin{abstract}
In Natural Language Processing (NLP), one of the most important tasks is text-to-SQL semantic parsing, which focuses on enabling users to interact with the database in a more natural manner. In recent years, text-to-SQL has made significant progress, but most were English-centric. In this paper, we introduce Ar-Spider \footnote{Our dataset is available at https://github.com/sasmohaimeed/Ar-Spider}, the first Arabic cross-domain text-to-SQL dataset. Due to the unique nature of the language, two major challenges have been encountered, namely schema linguistic and SQL structural challenges. In order to handle these issues and conduct the experiments, we adopt two baseline models  LGESQL \cite{cao2021lgesql} and S\textsuperscript{2}SQL \cite{hui2022s}, both of which are tested with two cross-lingual models to alleviate the effects of schema linguistic and SQL structure linking challenges. The baselines demonstrate decent single-language performance on our Arabic text-to-SQL dataset, Ar-Spider, achieving 62.48\% for S\textsuperscript{2}SQL and 65.57\% for LGESQL, only 8.79\% below the highest results achieved by the baselines when trained in English dataset. To achieve better performance on Arabic text-to-SQL, we propose the context similarity relationship (CSR) approach, which results in a significant increase in the overall performance of about 1.52\% for S\textsuperscript{2}SQL and 1.06\% for LGESQL and closes the gap between Arabic and English languages to 7.73\%.
\end{abstract}



%\keywords{ACM proceedings, \LaTeX, text tagging}

\maketitle

\section{Introduction}

The semantic parsing task aims to transform a natural language (NL) sentence into a formal representation of its meaning, which could be a logical form or structured query language (SQL). This task has various applications, including text-to-SQL \cite{qin2022survey}, task-oriented dialogue \cite{zhang2020recent}, and code generation \cite{dehaerne2022code}. In this paper, we focus on text-to-SQL, where numerous applications and benchmarks have been developed in English language. However, only a few are designed for non-English languages \cite{min2019pilot}\cite{nguyen2020pilot}\cite{jose2021mrat}\cite{bakshandaeva2022pauq}\cite{dou2023multispider}, and none are designed for Arabic speakers. Since SQL is a universal semantic representation, it is worthwhile to study it in languages other than English. However, different languages may pose different challenges to the semantic parsing models. Considering the Arabic language for example, there are two grand challenges that need to be addressed: (1) Schema linguistic challenge: the schema tables and columns of the relational databases are represented in English, which makes it more challenging to map NL words in Arabic into database entities in English. (2) SQL structural challenge: the SQL programing language increases the difficulty of mapping NL words written in Arabic letters to SQL clauses and operators. 

To extend the text-to-SQL into cross-language domain and address the two aforementioned challenges, we design Ar-Spider, the first Arabic cross-domain Text-to-SQL dataset for the semantic parsing task. Specifically, we create this Arabic version by manually translating the English Spider \cite{yu2018spider} dataset into Arabic. The dataset consists of 9691 questions with corresponding SQL queries over 166 databases, translated by two qualified translators and verified by three professional graduate computer science students. 

We conducted an empirical evaluation by using two state-of-the-art baseline models, LGESQL \cite{cao2021lgesql} and S\textsuperscript{2}SQL \cite{hui2022s}. We extend the models with two cross-lingual pre-training language encoders mBERT \cite{DBLP:journals/corr/abs-1810-04805}, and XLM-R \cite{DBLP:journals/corr/abs-1911-02116}. The cross-lingual models mitigate the effects of schema linguistic and partially handle SQL structural challenges. 

To further improve the performance, we introduce a novel approach, which establishes \textbf{C}ontext \textbf{S}imilarity \textbf{R}elationships (CSR) to the question-schema graph encoder in such a way that words from different languages are mapped to one another based on their context similarity in the embedding space.

Regardless of whether the inputs are encoded using mBERT or XLM-R cross-lingual models, our CSR approach improves LGESQL \cite{cao2021lgesql} and S\textsuperscript{2}SQL \cite{hui2022s} models. Specifically, a performance of 66.63\% was achieved by LGESQL + XLM-R + CSR, outperforming the highest baseline of 65.57\% achieved by LGESQL + XLM-R. Furthermore, LGESQL + mBERT + CSR, S\textsuperscript{2}SQL + mBERT + CSR, and S\textsuperscript{2}SQL + XLM-R + CSR all produced better results than those without CSR approach by 0.10\%, 2.33\%, and 1.52\%, respectively.

The contributions of this paper are as follows: 
\begin{enumerate}
    \item To the best of our knowledge, our Ar-Spider is the first Arabic cross-domain text-to-SQL dataset.
    \item A series of experiments are conducted on the Ar-Spider using two state-of-the-art models: LGESQL \cite{cao2021lgesql} and S\textsuperscript{2}SQL \cite{hui2022s}, both evaluated with two pre-training cross-lingual language models: mBERT \cite{DBLP:journals/corr/abs-1810-04805} and XLM-R \cite{DBLP:journals/corr/abs-1911-02116}.
    \item We propose a {C}ontext \textbf{S}imilarity \textbf{R}elationship (CSR) approach to mitigate the effect of the schema linguistic problem by injecting new relationship into the question-schema graph encoder, which increases the performance of the models LGESQL \cite{cao2021lgesql} and S\textsuperscript{2}SQL \cite{hui2022s}.
\end{enumerate}

\section{Related Work}
The Text-to-SQL semantic parsing has been extensively studied, especially with English datasets such as WikiSQL \cite{zhong2017seq2sql} and Spider \cite{yu2018spider}. Despite Spider \cite{yu2018spider} having fewer questions than WikiSQL \cite{zhong2017seq2sql}, it covers more complex SQL queries, answers longer questions, and has more tables and databases across different domains. Additionally, the training and testing set of the Spider \cite{yu2018spider} uses different domains, which ensures model generalization. As a further extension to the Spider \cite{yu2018spider} dataset, SparC \cite{yu2019sparc} and CoSQL \cite{yu2019cosql} have been released for dialogue-based Text-to-SQL, in which a series of interrelated questions in a single context are handled.

Several attempts have been made to repurpose Spider \cite{yu2018spider} dataset in other languages. The first attempt is CSpider \cite{min2019pilot}, a Chinese version of Spider \cite{yu2018spider}. There have been two datasets used in their experiment, one where Spider \cite{yu2018spider} questions have been translated by professional human translators and the other with machine translation. Results indicate that human translation has a significant advantage over machine translation. Furthermore, the model was tested using two word embedding representations. Cross-lingual embedding is found to be more effective than monolingual embedding, possibly because schema tables and columns are not translated, thus cross-lingual embedding provides a better connection between natural language questions and database columns. Their findings show that professional translators are necessary to adapt English datasets into another language and to test different cross-lingual embedding models with the new dataset.

Three additional versions of the Spider \cite{yu2018spider} have been made, in Vietnamese \cite{nguyen2020pilot}, Portuguese \cite{jose2021mrat}, and Russian \cite{bakshandaeva2022pauq}. The Vietnamese version \cite{nguyen2020pilot} include translated questions, database tables and columns, which help their experiment on the monolingual embedding in Vietnamese PhoBERT \cite{nguyen2020phobert} to outperform the cross-lingual embedding XLM-R \cite{DBLP:journals/corr/abs-1911-02116}. 

The Portuguese \cite{jose2021mrat} version translates only Spider \cite{yu2018spider} questions, and demonstrates the usefulness of cross-lingual models such as the m-BART-50 \cite{tang2020multilingual} when dealing with languages other than English. It has also been reported that training the model with both English and Portuguese questions at the same time may improve its performance.

In the Russian \cite{bakshandaeva2022pauq} version, Spider questions were translated only without translating the columns and tables of the database. Similar to CSpider\cite{min2019pilot}, both human translation dataset as well as machine translation dataset are provided. The results indicate that human translation is significantly superior to machine translation. Moreover, similar to the Portuguese finding, combining both Russian and English questions during the training has improved the model performance in two baselines.

The largest attempt was done by \cite{dou2023multispider} where they built a dataset that contains seven languages (English, Chinese, Vietnamese, German, French, Spanish, and Japanese). The questions for the first three languages mentioned are taken from existing datasets \cite{yu2018spider}, \cite{min2019pilot}, \cite{nguyen2020pilot}, respectively. Results indicate that training with all seven languages then testing in one language is more effective than training and testing in only one language.

\begin{table}[]
\caption{Example questions with corresponding SQL queries in Ar-Spider.}
\label{tab:sample_of_dataset}
\renewcommand{\arraystretch}{1.3}
\begin{tabular}{l}
\hline
\textbf{Sample 1: applying only one table in one database.}                                                                                                         \\ \hline
\multicolumn{1}{c}{\textbf{SQL Query}}                                                                                                                              \\
SELECT count(*) FROM products                                                                                                                                       \\
\multicolumn{1}{c}{\textbf{English Question}}                                                                                                                       \\
Count the number of products.                                                                                                                                       \\
\multicolumn{1}{c}{\textbf{Translated Arabic Question}}                                                                                                             \\
\multicolumn{1}{l}{\arabictext{احسب عدد المنتجات.}}                                                                                                                              \\
                                                                                                                                                                    \\ \hline
\textbf{Sample 2: applying multiple tables in one database.}                                                                                                        \\ \hline
\multicolumn{1}{l}{\textbf{SQL Query}}                                                                                                                              \\
\begin{tabular}[l]{@{}l@{}}SELECT count(*) , T2.name FROM products AS T1 \\ JOIN  manufacturers AS T2 ON T1.Manufacturer = \\ T2.code GROUP BY T2.name\end{tabular} \\
\multicolumn{1}{l}{\textbf{English Question}}                                                                                                                       \\
How many products are there for each manufacturer?                                                                                                                  \\
\multicolumn{1}{l}{\textbf{Translated Arabic Question}}                                                                                                             \\
\multicolumn{1}{l}{\arabictext{كم من المنتجات لكل شركة صناعية؟}}                                                                                                                 \\ \hline
\end{tabular}
\end{table}

\section{Dataset}
The Spider \cite{yu2018spider} dataset has 10,181 questions separated as training, development, and testing sets. Since the authors did not make the testing set publicly available, we have translated only training and development sets of 9691 questions and ensured that no databases overlap between the two sets. For a more efficient translation, we use GPT 3 \cite{brown2020language} to translate the Spider questions into Arabic, then ask professional translators to post-edit them individually. In Section 3.2, we demonstrate how this method speeds up the translation process and increases the diversity of question words. Specifically, two native Arabic speakers specializing in English translation services were involved in such a post-editing, each of them was assigned with 83 databases. In the following step, three graduate computer science students, who are proficient in SQL, read the original English version of the questions, ensure that the post-edit Arabic translation is accurate, and then verify that the Arabic questions are consistent with the SQL query. A sample of the dataset is shown in \autoref{tab:sample_of_dataset}.

\subsection{Dataset Statistics}
Multiple attempts have been made to adapt the English Spider \cite{yu2018spider} dataset to other languages \cite{min2019pilot, nguyen2020pilot, jose2021mrat, bakshandaeva2022pauq, dou2023multispider}. Each of them consists of the same number of questions, SQL queries, and database schemas. The statistics and splitting settings of the Ar-Spider are shown in \autoref{tab:stat}.

\begin{table}[]
\caption{A comparison between the English Spider dataset, and the Arabic version Ar-Spider. \#Q denotes the number of questions, \#SQL donates the number of distinct SQL queries,  \#DB donates the number of databases, and \#Tables/DB represents the average number of tables per database.}
\label{tab:stat}
\renewcommand{\arraystretch}{1.3}
\setlength{\tabcolsep}{6.5pt}
\begin{tabular}{cc|cccc}
\cline{3-6}
                                                      &                & \textbf{\#Q} & \textbf{\#SQL} & \textbf{\#DB} & \textbf{\#Tables/DB} \\ \hline
\multicolumn{1}{c|}{\textbf{English}}                 & \textbf{all}   & 10181        & 5693           & 200           & 5.1                  \\ \hline
\multicolumn{1}{c|}{{\textbf{Arabic}}} & \textbf{all}   & 9691         & 5277           & 166           & 5.28                 \\
\multicolumn{1}{c|}{}                                 & \textbf{train} & 8657         & 4714           & 146            & 5.45                  \\
\multicolumn{1}{c|}{}                                 & \textbf{test}  & 1034         & 563           & 20            & 4.05                 \\ \hline
\end{tabular}
\end{table}

\subsection{Using GPT3 as Translation Model}

At the beginning of the translation process, 600 English questions were translated manually without the assistance of any machine translation tools. As a result of following this process, we find that there are two disadvantages. As shown in \autoref{tab:gpt_translation}, it takes approximately 70 seconds per question to read an English question, translate it into Arabic, and ensure that it aligns with the SQL query. The completion of all 600 questions takes around 11 hours and 42 minutes. Therefore, it will need 188 hours to complete the translation process of the 9691 questions. Alternatively, we replaced the manual translation with machine translation and post-edited the questions. In this way, the process has sped up to 30 seconds per question. Based on our final estimation, it took approximately 80 hours to complete the 9691 questions.

\begin{table}[]
\caption{Comparison of the time to complete the translation process with and without using GPT3 as a machine translation tool. Q refers to question. Sec refers to seconds}
\label{tab:gpt_translation}
\renewcommand{\arraystretch}{1.3}
\setlength{\tabcolsep}{4pt}
\begin{tabular}{c|cccc}
\hline
\textbf{\begin{tabular}[c]{@{}c@{}}Translation \\ process\end{tabular}} & \textbf{Sec / Q} & \textbf{600 Q} & \textbf{9691 Q} & \textbf{\begin{tabular}[c]{@{}c@{}}Word\\  diversity\end{tabular}} \\ \hline
\textbf{With GPT3}                                                      & 30               & 5 hours        & 80.1 hours      & \checkmark                                                                  \\
\textbf{Without GPT3}                                                   & 70               & 11.7 hour      & 188 hours       & X                                                                 \\ \hline
\end{tabular}
\end{table}

\begin{figure*}[h]
  \centering
  \includegraphics[width=\textwidth]{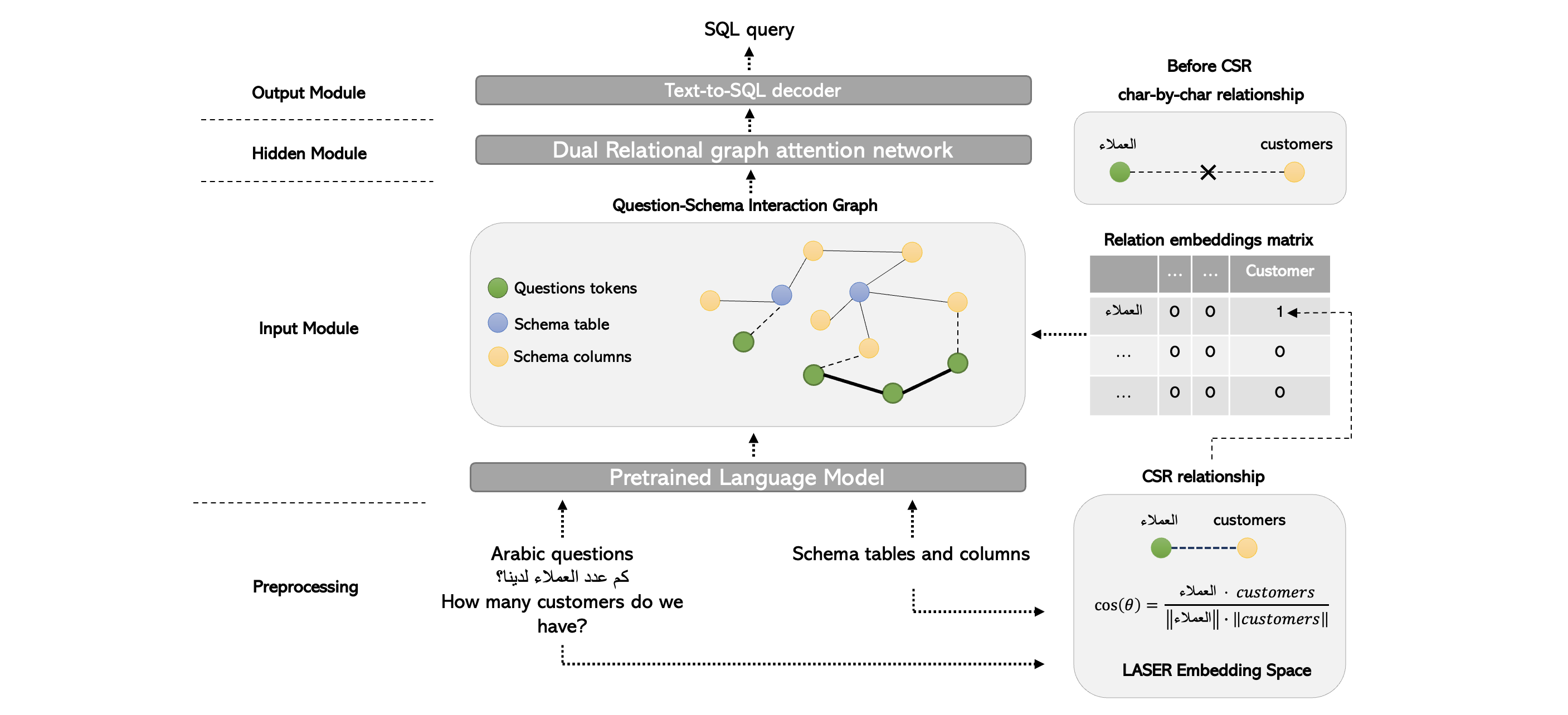}
  \caption{An illustration of the overall model architecture. There are three types of relations between nodes in the graph. Dot-line represents question-table-cosine-matches or question-column-cosine-matches. Bold-line represents question structure relationships between the nodes of the question tokens. Straight-line indicates schema structure relationships, such as primary and foreign keys.  In the figure, the top right subgraph shows how question schema relations were created before CSR.}
  \label{fig:org4}
\end{figure*}

Secondly, despite the fact that two translators have worked on these 600 questions, the questions suffer from the diversity of Arabic words. For example, when translating the sentence "show students name" into Arabic, we always translate "show" as "\arabictext{ أعرض}". However, there are several other translations that can be used, such as "\arabictext{ أظهر}", and "\arabictext{ أوجد}". For this reason, we have used the GPT3 \cite{brown2020language} model to translate the questions into Arabic and allowed the translators to make any necessary modifications to the questions after translation. A comparison of our translation process with and without the machine translation model is presented in \autoref{tab:gpt_translation}. Moreover, we did not utilize Google NMT \cite{wu2016googles} for the translation process because GPT3 provides a greater range of words diversity.

In addition, it is necessary to address the question of why the Arabic questions are not simply translated into English prior to the queries being generated without the involvement of a verifiers. GPT3 is a very good translation tool, but there are some questions that it does not ask in the same manner as a native Arabic speaker and there are others for which the context of the question is not recognized.  By involving a verifiers who is a native Arabic speakers, we ensure that the essence of the question is preserved and accurately represented in the SQL query. In addition, based on previous research \cite{min2019pilot} \cite{bakshandaeva2022pauq}, we can conclude that manual translation is more effective than using a machine translation tool.

\section{Models}
When our empirical investigation was conducted, there were over 75 models in the leaderboard of Spider \cite{yu2018spider}. To select baselines, the following three criteria were considered:
\begin{enumerate}
    \item Some state-of-the-art models did not publish their source code. As a result, these models were not considered during our experiments.
    \item For better model reusability, it is important to choose models that take into account only the exact match accuracy metrics that are used in our experiment and do not consider other implementations of unrelated metrics.
    \item It is necessary to integrate other pre-training language models without modifying the overall model structure. Therefore, baselines must be composed of partially independent encoders, architectures, and decoding processes.
    
\end{enumerate}

Based on the aforementioned criteria, we have selected LGESQL \cite{cao2021lgesql} and S\textsuperscript{2}SQL \cite{hui2022s} as the most suitable Spider models for this experiment.
Both the LGESQL and S\textsuperscript{2}SQL models have a similar encoder-decoder architecture, which consists of three parts, namely input graph, hidden line graph, and output graph modules. S\textsuperscript{2}SQL has added more implementations to the input graph modules in the English Spider \cite{yu2018spider} version, resulting in a slight increase in performance with the ELECTRA \cite{clark2020electra} pre-training language model. However, It performs less well when using cross-lingual pre-training models, such as mBERT \cite{DBLP:journals/corr/abs-1810-04805} and XLM-R \cite{DBLP:journals/corr/abs-1911-02116}.

During the preprocessing phase, both LGESQL and S\textsuperscript{2}SQL models generate relationships between the graph nodes and store them in relation embedding matrix. LGESQL handles two types of relation structures: \textbf{Linking Structure} that connects the tokens in a question to their counterparts in the database schema, and \textbf{Schema Structure} that connects the relationships between database schema items (like primary-foreign keys). On the other hand, S\textsuperscript{2}SQL captures the two aforementioned relations as well as the \textbf{Question Structure}, which represents the syntactic dependencies between question tokens. In other words, it provides an indication of which words depend on which other words to form meaningful sentences.

\subsection{Input Graph Module}

Initial embedding of nodes and edges of the graph are provided by the input module, in which the nodes represent question tokens, schema tables, and schema columns, while the edges represent relationships between nodes. For the LGESQL model, node embedding are represented using either word embedding Glove \cite{pennington2014glove} or pre-training language models such as BERT \cite{devlin2018bert} and ELECTRA \cite{clark2020electra}. Meanwhile, S\textsuperscript{2}SQL uses only the best-performing pre-training language model ELECTRA. On the other side, the relation edge embedding are retrieved directly from a parameter matrix.   
\vspace{0.1cm}
\subsection{Hidden Line Graph Module}
\vspace{0.2cm}
The hidden line graph module will capture the relational structure between the initially generated node embedding using a relational graph attention network. The node embedding are updated in iterative processes, in which nodes gather and aggregate information from their neighboring nodes based on the attention mechanism. In addition, there is an enhanced line graph that represents the edges between the nodes in the original graph, which allows the model to consider both the importance of the nodes and their edges. Due to the fact that some embedding for different edges might become too similar or coupled during optimization, S\textsuperscript{2}SQL \cite{hui2022s} introduces an orthogonality condition that will add a penalty if the embedding become too similar.
\vspace{0.1cm}
\subsection{Output Graph Modules}
\vspace{0.2cm}
Using a grammar-based syntactic neural decoder, they construct the abstract syntax tree (AST) of the predicted query in depth-first search order. At each decoding step, the output is either an APPLYRULE action that expands the current partially generated AST, or 2) SELECTTABLE or SELECTCOLUMN actions that select one schema element. In addition, they performed an auxiliary task known as graph pruning, which will assist in distinguishing between relevant and irrelevant schema items.

As a brief summary of the differences between LGESQL and S\textsuperscript{2}SQL, encoding S\textsuperscript{2}SQL integrates syntax dependency among question tokens into the relational graph attention network and introduces a decoupling constraint for generating diverse relation embedding.
% \usepackage{booktabs}
\begin{table*}[h!]
\caption{A comparison that shows analysis of different cosine similarity functions results derived from four cross-lingual models. The Arabic words \arabictext{ملك} = king, \arabictext{سفر} = travel, \arabictext{رجل} = man, \arabictext{رقم الهاتف} = phone number, \arabictext{هاتف} = phone}
\label{tab:laser1}
\begin{tabular}{@{}c|cccccc@{}}
\toprule
 & \textbf{King \arabictext{ملك}} & \textbf{Travel \arabictext{سفر}} & \textbf{Man \arabictext{سفر}} & \textbf{Phone number \arabictext{رجل}} & \textbf{Phone number \arabictext{رقم الهاتف}} & \textbf{Phone number \arabictext{هاتف}} \\ \midrule
\textbf{mBERT} & 77.73 & 54.50 & 73.58 & 82.48 & 82.26 & 71.89 \\
\textbf{XLM}   & 68.76 & 78.45 & 70.48 & 65.75 & 61.53 & 58.90 \\
\textbf{SBERT} & 83.53 & 85.99 & 62.64 & 25.29 & 91.91 & 69.51 \\
\textbf{LASER} & 86.37 & 88.44 & 65.06 & 61.04 & 85.16 & 84.49 \\ \bottomrule
\end{tabular}
\end{table*}

\section{CSR Approach}
\vspace{0.2cm}
Schema linking represents one of the major challenges we face in the text-to-SQL task, which involves mapping natural language words into their corresponding schema tables and columns. As shown in the Question-Schema Interaction Graph in Figure \ref{fig:org4}, the model generates an initial graph that consists of three types of nodes: questions tokens, schema tables, and schema columns. Furthermore, the model will generate many relationships, including linking structure, schema structure, and question structure relationships. For the linking structure relationships in the original English spider \cite{yu2018spider} dataset, the model will compare each question token character-by-character with each table. When they are exactly similar, as for instance the question token ``customers" and the table ``customers", the LGESQL \cite{cao2021lgesql} creates a relationship edge between them and name it ``question-table-exact-match". In addition, if they partially match, for example, the question token ``store" and the table ``store name", LGESQL will create a relationship edge between them called question-table-partial-match. Furthermore, the same procedure will be applied to each question token with each schema column.

For Ar-Spider, a new challenge arises, namely the schema linguistics challenge, in which the question tokens are written in Arabic, while the schema tables and columns are written in English. Thus, a character-by-character comparison of question tokens with schema tables or schema columns will always result in a no-match relationship. To address this issue, we propose a new matching approach called Context Similarity Relationship (CSR). Rather than comparing character by character, we will calculate the cosine similarity between the question tokens and the schema table or column in the embedding space. Those cosine similarity that exceed a certain threshold will be taken into consideration. In our case, the threshold is 78\%.

To implement our CSR approach, during preprocessing, we utilized the Language-Agnostic Sentence Representations (LASER) \cite{artetxe2019massively} model. The reason we selected LASER over other cross-lingual models is that LASER maps a sentence in any language to a point in a high-dimensional embedding space in such a way that the same statement in any language will end up in the same neighborhood. Therefore, the vector representation of an English sentence would be similar to the vector representation of its Arabic translation. Consequently, words with the same meaning, but in different languages, will be located in the same neighborhood.

As shown in \autoref{fig:org4}, before the question tokens and schema tables and columns are encoded using a pre-trained language model, LASER \cite{artetxe2019massively} will be used to create the relation embedding matrix. This is done by comparing every Arabic question token to every table and column. It will be determined that there is a relationship if the cosine similarity is at least 78\%. It can be seen from the upper right corner of Figure \ref{fig:org4} that before the CSR approach, Arabic tokens would have no relation to schema tables and columns, despite their similar meanings. 

For better insight, we conducted some analysis to determine how many relationships were established per sample, and the CSR approach showed that the relation embedding matrix showed more relationships with an average of 0.8 question-table-cosine-match and 2.8 question-column-cosine-matches. One might argue that adding these relationships would add additional complexity to the model, which could negatively affect its efficiency. However, without the CSR approach, the relationship matrix with the question tokens and schema items relationships is already built in with zeros in all of the relationships. In contrast, our approach allows the relationship matrix to carry more information, which results in better performance across all baselines.

\subsection{Ablation Study: Compare LASER with Other Models}
Our CSR approach was specifically designed using LASER, and we have done an ablation study to demonstrate the reasons for our decision.  A comparison of LASER with three cross-lingual models is given in \autoref{tab:laser1}, including mBERT \cite{DBLP:journals/corr/abs-1810-04805}, XLM \cite{lample2019cross}, and SBERT \cite{reimers2019sentence}. While mBERT and XLM were not developed with the intention of embedding sentences, LASER and SBERT have been designed for the purpose of embedding sentences. Based on the results, it can be seen that mBERT and XLM have a high cosine similarity percentage of unrelated words and a low percentage of related words. For LASER and SBERT, this is not the case. Always similar words will have a high cosine similarity percentage, while unrelated words will have a lower percentage. Even though SBERT \cite{reimers2019sentence} performs well, if we compare a question token with schema items where the words are partially similar as shown in the last column of \autoref{tab:laser1}, SBERT will not perform well. Due to this, we have utilized LASER in our experiment.

\section{Experiments}
\subsection{Experimental Setup}
\subsubsection{Evaluation Metric}
Results are reported for all examples using the exact match accuracy metric, which is the same metric used in Spider \cite{yu2018spider}. It measures whether the SQL clauses, tables, and columns of the predicted SQL query are equivalent to the gold SQL query. Furthermore, \cite{yu2018spider} has divided the SQL queries into four difficulty levels (easy, medium, hard, and extra hard). In general, the more SQL keywords in a SQL query, the harder the query is considered to be.
\subsubsection{Baselines and Pre-trained Encoder models}
With LGESQL \cite{cao2021lgesql} and S\textsuperscript{2}SQL \cite{hui2022s} baseline models, we selected two pre-trained cross-lingual encoders, mBERT \cite{DBLP:journals/corr/abs-1810-04805} and XLM-Roberta-Large \cite{DBLP:journals/corr/abs-1911-02116}. The structure of mBERT is similar to that of BERT \cite{devlin2018bert}. The only difference is that mBERT has been pre-trained on concatenated Wikipedia data for 104 languages, including Arabic. As for XLM-Roberta-Large, it has also been pre-trained on text in 100 languages, including Arabic, and reported state-of-the-art results in a number of Arabic natural language processing tasks \cite{khalifa2021self} \cite{el2022adasl} \cite{lan2020empirical}.

\subsection{Experimental Results}

In our experiments, we want to answer the following research questions (RQs):
\begin{itemize}
    \item  RQ1 What are the performance differences between models that were trained in the English Spider \cite{yu2018spider} dataset and those that were trained on the Arabic Ar-spider dataset?
    \item RQ2 How does the performance of different cross-language models differ?
    \item RQ3 Does combining the English and Arabic versions of the datasets enhance the models performance?
    \item RQ4 By using our CSR approach, what impact does it have on the performance of the models?
\end{itemize}


\begin{table}[]
\caption{Exact match accuracy results of LGESQL and S\textsuperscript{2}SQL models over the original English Spider dataset.}
\label{tab:Original-Spider}
\begin{tabular}{@{}c|ccc@{}}
\toprule
\textbf{Dataset}        & \textbf{Model}          & \textbf{Encoder} & \textbf{Accuracy} \\ \midrule
\multirow{4}{*}{Spider} & \multirow{2}{*}{LGESQL} & mBERT            & 72.79             \\
                        &                         & XLM-R            & 74.36             \\ \cmidrule(l){2-4} 
                        & \multirow{2}{*}{S\textsuperscript{2}SQL}  & mBERT            & 67.79             \\
                        &                         & XLM-R            & 70.41             \\ \bottomrule
\end{tabular}
\end{table}

Based on three different datasets, we evaluated the performance of LGESQL \cite{cao2021lgesql} and S\textsuperscript{2}SQL \cite{hui2022s} models. \autoref{tab:Original-Spider} shows the results of the models that were trained on the original English Spider \cite{yu2018spider} dataset. The second dataset is Ar-Spider and \autoref{tab:Arabic-version} shows its related results. \textbf{RQ3} is addressed by the third dataset, which is a combination of both Spider \cite{yu2018spider} and Ar-spider datasets, and their corresponding results are displayed in \autoref{tab:english_and_arabic}.

\subsubsection{Spider Dataset}
In \autoref{tab:Original-Spider}, we tested the LGESQL and S\textsuperscript{2}SQL models using the original Spider \cite{yu2018spider} dataset, without modifying the models except in one respect, which was integrating the cross-lingual models mBERT and XLM-R with the baselines. The Spider \cite{yu2018spider} dataset is available at https://yale-lily.github.io/spider. Despite the fact that the S\textsuperscript{2}SQL model is more efficient with the ELECTRA \cite{clark2020electra} pre-trained language model, it does not perform well with any of the cross-lingual models. Compared to S\textsuperscript{2}SQL + mBERT, LGESQL + mBERT has an overall performance of 71.79\%, outperforming S\textsuperscript{2}SQL + mBERT by 1.96\%. Additionally, LGESQL with XLM-R achieved the highest results with 74.36\%, outperforming S\textsuperscript{2}SQL + XLM-R by 3.95\%.

% \usepackage{booktabs}
% \usepackage{multirow}
\begin{table}[]
\caption{Detailed comparison of the exact match results of LGESQL and S\textsuperscript{2}SQL over Ar-Spider with and without CSR.}
\label{tab:Arabic-version}
\begin{tabular}{@{}ccccc@{}}
\toprule
\textbf{Dataset}      & \textbf{Model}        & \textbf{Encoder}       & \textbf{CSR} & \textbf{Accuracy} \\ \midrule
\multicolumn{1}{c|}{\multirow{8}{*}{Ar-Spider}} & \multicolumn{1}{c|}{\multirow{4}{*}{LGESQL}} & \multirow{2}{*}{mBERT} & - & 55.32 \\
\multicolumn{1}{c|}{} & \multicolumn{1}{c|}{} &                        & Yes          & 55.42             \\ \cmidrule(l){3-5} 
\multicolumn{1}{c|}{} & \multicolumn{1}{c|}{} & \multirow{2}{*}{XLM-R} & -            & 65.57             \\
\multicolumn{1}{c|}{} & \multicolumn{1}{c|}{} &                        & Yes          & 66.63             \\ \cmidrule(l){2-5} 
\multicolumn{1}{c|}{}                           & \multicolumn{1}{c|}{\multirow{4}{*}{S\textsuperscript{2}SQL}}  & \multirow{2}{*}{mBERT} & - & 52.51 \\
\multicolumn{1}{c|}{} & \multicolumn{1}{c|}{} &                        & Yes          & 54.84             \\ \cmidrule(l){3-5} 
\multicolumn{1}{c|}{} & \multicolumn{1}{c|}{} & \multirow{2}{*}{XLM-R} & -            & 62.48             \\
\multicolumn{1}{c|}{} & \multicolumn{1}{c|}{} &                        & Yes          & 64                \\ \bottomrule
\end{tabular}
\end{table}

\subsubsection{Ar-Spider Dataset}
In \autoref{tab:Arabic-version}, the models were trained on the Ar-spider dataset. As for the LGESQL model, we did not modify the model except for tokenizing the Arabic questions using the Stanza \cite{el2022adasl} Arabic version library and integrating encoders mBERT \cite{DBLP:journals/corr/abs-1810-04805} and XLM-R \cite{DBLP:journals/corr/abs-1911-02116}. As for S\textsuperscript{2}SQL, we have made the same modification as to the question dependency parser to accommodate Arabic questions.  According to the accuracy results \autoref{tab:Original-Spider}, baseline models without CSR approach are ranked ascendingly. LGESQL + mBERT produces the lowest results of 52.51\%, while LGESQL + XLM-R produces the highest results of 65.57\%. S\textsuperscript{2}SQL performs less than LGESQL because it captures dependency relationships in the question grammatical structure, requiring an encoder that is rich in vocabulary in the target language in our case Arabic. While both mBERT and XLM-R have a good amount of Arabic vocabulary, they are still not enough.

Furthermore, to answer question \textbf{RQ4}, through the use of our CSR context similarity approach, three of the baselines have shown significant improvements of 2.33\%, 1.52\%, and 1.06\% for S\textsuperscript{2}SQL + mbert, S\textsuperscript{2}SQL + XLM-R, and LGESQL + XLM-R, respectively. However, the model has slightly improved by 0.1\% for LGESQL + mBERT. This success of CSR can be attributed to the fact that prior to CSR, all linking structures relationships between question tokens and schema entities were never matched, since the question was in Arabic and the schema entities were in English. In contrast, using CSR, more than 14000 relationships have been established with a similarity threshold of 78\%, leading to an increase in performance across all baselines.

For \textbf{RQ1}, the models are expected to perform less well with the Arabic Ar-Spider dataset. It is due to two challenges: the schema linguistic challenge and the SQL structure challenge. Nevertheless, LGESQL + XLM-R + CSR obtains good results, achieving 66.63\%, only 7.73\% less than the best results obtained on the English Spider \cite{yu2018spider} dataset. For \textbf{RQ2}, the results indicated that the use of a large cross-lingual model, such as XLM-R \cite{DBLP:journals/corr/abs-1911-02116}, is crucial for capturing context effectively and achieving better performance.

% \usepackage{booktabs}
% \usepackage{multirow}
\begin{table}[]
\caption{Exact accuracy results of LGESQL model by combining both English Spider and Arabic Ar-Spider datasets.}
\label{tab:english_and_arabic}
\begin{tabular}{@{}cccc@{}}
\toprule
\textbf{Dataset}                                         & \textbf{Model}          & \textbf{Encoder} & \textbf{Accuracy} \\ \midrule
\multicolumn{1}{c|}{\multirow{2}{*}{Spider + Ar-Spider}} & \multirow{2}{*}{LGESQL} & mBERT            & 52.03             \\
\multicolumn{1}{c|}{}                                    &                         & XLM-R            & 64.31                 \\ \bottomrule
\end{tabular}
\end{table}

\subsubsection{Spider and Ar-Spider.}
Based on \cite{jose2021mrat} \cite{bakshandaeva2022pauq} \cite{dou2023multispider}, it was found that training the models both with English spider \cite{yu2018spider} dataset and their target language resulted in a significant improvement in performance. To answer \textbf{RQ3}, \autoref{tab:english_and_arabic} indicates that this is not the case for our experiment. We have noticed a slight decrease in performance when we combine Spider \cite{yu2018spider} with Ar-Spider, then test the model in the Ar-Spider testing set. This could be due to the fact that Arabic has a more complex morphology, meaning its word forms undergo a variety of changes to convey different grammatical functions. In Arabic, nouns take on different forms depending on their grammatical case, number, and gender. Thus, combining Arabic with similar languages such as Persian or Urdu may enhance the performance.

\section{Conclusions}
This paper presents the first Arabic cross-domain text-to-SQL dataset named Ar-Spider, revised from the original English Spider dataset \cite{yu2018spider}. In order to ensure high quality, we translated Spider by two professional translators and had it verified by three  computer science graduate students. Dataset was evaluated using two strong baseline models, LGESQL \cite{cao2021lgesql} and S\textsuperscript{2}SQL \cite{hui2022s}, demonstrating that a larger cross-lingual model will significantly improve the performance. Additionally, we identify some challenges related to the Arabic language, Scehma linguistics, and SQL structure challenges. Along with using cross-lingual models to minimize the effect of these challenges, we introduce the Context Similarity Relationships approach that has resulted in a notable improvement on performance for all baselines, where the highest performance is 66.63\% achieved by LGESQL + XLM-R + CSR. Further, we discuss the reasons for selecting LASER \cite{artetxe2019massively} as the model to compute the cosine similarity angle between question tokens and schema items. Additionally, we have shown that combining English and Arabic datasets during training has not improved the performance. In future work, it would be useful to develop better methods for linking Arabic questions to English schema items, such as utilizing large language models or by fine-tuning pre-trained models specifically for the arabic-english schema linking task.

\bibliographystyle{ACM-Reference-Format}
\bibliography{main}

\end{document}